\pdfoutput=1

\documentclass[11pt]{article}
\usepackage{paralist}
\usepackage{booktabs}
\usepackage[preprint]{acl}

\usepackage{times}
\usepackage{latexsym}
\usepackage[T1]{fontenc}

\usepackage[utf8]{inputenc}
\usepackage[shortlabels]{enumitem}

\usepackage{microtype}
\usepackage{multirow}
\usepackage{makecell}

\usepackage{amsmath,amssymb,amsfonts}
\usepackage{algorithmic}
\usepackage[linesnumbered,ruled,vlined]{algorithm2e}
\usepackage{graphicx}
\usepackage{textcomp}
\usepackage{xcolor}
\usepackage{subcaption}
\usepackage{multirow}
\usepackage{booktabs}
\usepackage{amsthm}
\usepackage{enumitem}
\usepackage{array}
\usepackage{stfloats}
\usepackage{url}
\usepackage{verbatim}
\usepackage{tabularx}
\usepackage{natbib}
\usepackage[utf8]{inputenc}
\usepackage{float}
\usepackage{hyperref}  
\usepackage{cleveref}

\usepackage{inconsolata}

\usepackage{graphicx}
\usepackage{amsmath}
\usepackage{array}
\usepackage{makecell}

\usepackage{tikz}
\usepackage{forest}
\useforestlibrary{edges}
\usetikzlibrary{shapes, arrows.meta, positioning}

\usepackage{xcolor}

\definecolor{softgreen}{HTML}{F4FBF3}   
\definecolor{softblue}{HTML}{F2F9FF}    
\definecolor{softyellow}{HTML}{FFFBEB}  
\definecolor{softgray}{HTML}{F7F8FA}    
\definecolor{softpurple}{HTML}{F7F2FC}  

\definecolor{myyellow}{HTML}{FFFBEB}    
\definecolor{mypurple}{HTML}{F7F2FC}    

\tikzset{
  leaf/.style={
    draw=black!40,
    rounded corners=2pt,
    line width=0.3pt,
    align=left,          
    inner sep=2pt,
    minimum width=25em,
    text width=25em,   
  },
  leaf1/.style={
    draw=black!40,
    rounded corners=2pt,
    line width=0.3pt,
    align=center,
    inner sep=2pt,
  },
  leaf3/.style={
    draw=black!40,
    rounded corners=2pt,
    line width=0.3pt,
    align=center,
    inner sep=2pt,
  },
}

\title{A Survey on Collaborating Small and Large Language Models for Performance, Cost-effectiveness, Cloud-edge Privacy, and Trustworthiness}

\author{Fali Wang, Jihai Chen\thanks{~Work done as an intern at Penn State.}, Shuhua Yang, Ali Al-Lawati, Linli Tang \\
Pennsylvania State University \\
University Park, USA \\
\AND 
Hui Liu \\
  Michigan State University \\
  East Lansing, USA \\
\And
Suhang Wang\thanks{~Corresponding author. Email: \texttt{szw494@psu.edu}} \\
Pennsylvania State University \\
University Park, USA \\
}

\newcommand{\textbfit}[1]{\textbf{\textit{#1}}}

\begin{document}
\maketitle
\begin{abstract}

Large language models (LLMs) have achieved remarkable progress across domains and applications, but face challenges such as high fine-tuning costs, inference latency, limited edge deployability, and reliability concerns. Small language models (SLMs), with compact, efficient, and adaptable features, offer promising solutions. Building on this potential, recent research explores collaborative frameworks that integrate their complementary strengths, leveraging SLMs’ specialization and efficiency with LLMs’ generalization and reasoning, to address diverse objectives across tasks and deployment scenarios. Motivated by these developments, this paper presents a systematic survey of SLM–LLM collaboration from the perspective of collaboration objectives. We propose a taxonomy covering four goals: \textit{performance enhancement, cost-effectiveness, cloud–edge privacy, and trustworthiness}. Under this framework, we review representative methods, summarize design paradigms, and outline open challenges and future directions toward efficient and secure SLM-LLM collaboration. The collected papers are available in \href{https://github.com/FairyFali/SLMs-Survey}{link here}.


\end{abstract}

\section{Introduction}

\begin{figure}[t]
    \centering
    \includegraphics[width=0.99\linewidth]{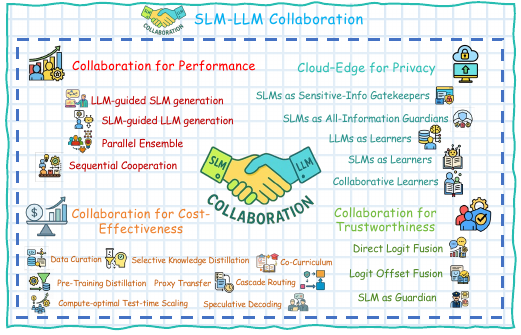}
    \caption{Detailed taxonomy of SLM–LLM collaboration, depicting the sub-taxonomy under the collaboration objectives of \textit{performance, cost-effectiveness, privacy, and trustworthiness.} }
    \label{fig:taxonomy2}
\end{figure}

Large language models (LLMs) have profoundly transformed multiple domains, including AI for science \cite{luo2022biogpt, allawati2025graphbasedmolecularincontextlearning, wang2024infuserki, wang2024self}, programming \cite{shi2024decoding}, and human-centered interaction \cite{zhang2024future}, primarily owing to their massive parameter scales. However, such scale also introduces several challenges: (1) fine-tuning is computationally intensive, limiting efficient model adaptation \cite{thawakar2025mobillama, liu2024mobilellm, xia2024understanding,zhang2024offline}; (2) large model size leads to inference latency, constraining real-time applications \cite{leviathan2023fast, kwon2023efficient, cai2024medusa}; (3) typical edge devices, such as mobile phones, personal computers, and small servers, lack the capacity to host these models, while cloud-based inference raises privacy and cost concerns \cite{carlini2021extracting, xu2024device, li2024large}; and (4) LLMs exhibit inherent reliability risks, including hallucinations and jailbreak vulnerabilities \cite{yao2024survey, farquhar2024detecting, wang2023decodingtrust}.
These challenges underscore the increasing need for customizable, cost-efficient, edge-deployable, and trustworthy AI solutions. Small language models (SLMs), characterized by their compact architecture, low computational cost, and adaptability, have emerged as promising counterparts to mitigate these issues, although their general reasoning and knowledge coverage remain more limited than those of LLMs. Consequently, harnessing the complementary strengths of SLMs and LLMs provides a viable pathway toward developing efficient, scalable, and reliable intelligent systems.

\begin{figure*}[h]
\centering
\begin{forest}
  for tree={
  forked edges,
  grow=east,
  reversed=true,
  anchor=center,
  parent anchor=east,
  child anchor=west,
  base=center,
  font=\scriptsize,
  rectangle,
  draw=black, 
  edge=black!50, 
  rounded corners,
  minimum width=2em,
  s sep=5pt,
  inner xsep=3pt,
  inner ysep=1pt
  },
  where level=1{text width=4.5em}{},
  where level=2{text width=6em,font=\scriptsize}{},
  where level=3{font=\scriptsize}{},
  where level=4{font=\scriptsize}{},
  where level=5{font=\scriptsize}{},
  [SLM-LLM Collaboration,rotate=90,anchor=north,inner xsep=8pt,inner ysep=3pt,edge=black!50,draw=black
    [Performance \\ \S \ref{sec:collaboration_performance}, edge=black!50, leaf3, fill=softgreen
      [Guidance-Generation \\ \S \ref{sec:advisory}, leaf1, fill=softgreen,
          [SynCID\cite{liang2024synergizing}{, }PiFi\cite{kim-etal-2025-plug}{, }SuperICL\cite{xu2023small}{, }Ensemble-SuperICL\cite{mojarradi2024improving}{, }LM-Guided CoT\cite{lee2024can}{, }CoWest\cite{jiao2024synergistic}{, }Synergistic Augmentation\cite{li-etal-2025-synergistic}{, }MSLLM\cite{qiao-etal-2025-mixture}{, }Flipping KD \cite{liFlippingKnowledgeDistillation2025}
            ,leaf,fill=softgreen]
      ]
      [Division-Fusion \\ \S \ref{sec:cospecialized_collaboration}, leaf1, fill=softgreen,
        [ELM\cite{gondara2025elm}{, }CaLM\cite{hsu2024calm}{, }ZeroNL2SQL\cite{fan2024combining}{, }KDSL\cite{chen2025knowledge}{, }Collab-RAG \cite{xu2025collab}{, }GCIE\cite{bao2024general}{, }HuggingGPT\cite{shen2023hugginggpt}{, }TrajAgent\cite{du2024trajagent},leaf,fill=softgreen]
      ]
    ]
    [Cost-Effectiveness
    \\ \S \ref{sec:collaboration_cost_effectiveness}, edge=black!50, leaf3, fill=myyellow,
      [Pre-Training Stage \\ \S \ref{sec:cost_effectiveness_pretrain}, leaf1, fill=myyellow,
        [SALT\cite{rawat2024little}{, }Perplexity-Based Data Pruning\cite{ankner2024perplexed}{, }LiGO\cite{wang2023learning}{, }Progressive Training\cite{yano2025efficient}{,}HyperCloning \cite{samragh2024scaling}{, }DistilBERT\cite{Sanh2019DistilBERT}{, }TinyBERT\cite{jiao2019tinybert}{, }MiniLM \cite{Wang2020MiniLM}{, }Pre-training Distillation\cite{peng2024pre}{, }DPT\cite{goyal2025distilled}{, }MINIPLM\cite{gu2024miniplm},leaf,fill=myyellow]
      ]
      [Fine-Tuning Stage \\ \S \ref{sec:cost_effectiveness_tuning}, leaf1, fill=myyellow,
        [LLKD\cite{li2024learning}{, }LLM-to-SLM KD\cite{jazbec2024efficient}{, }HAWKEYE\cite{she2025hawkeye}{, }Causal Distillation\cite{muhebwa2025causal}{, }C2KD\cite{chen2025c2kd}{, }Agent Distillation~\cite{kang2025distilling}{, }LSC4Rec\cite{lv2025collaboration}{, }SmallPlan\cite{pham2025smallplan}{, }LLM-to-SLM\cite{bergner2024think}{, }LoRAM\cite{zhang2025train}{, }LiteMOE\cite{zhuang2024litemoe}{, }GateKeeper\cite{rabanser2025gatekeeper}{, }Logit-fusion\cite{fan2025g},leaf,fill=myyellow]
      ]
      [Inference Stage \\ \S \ref{sec:cost_effectiveness_inference}, leaf1, fill=myyellow,
        [Frugal{GPT}\cite{chen2024frugalgpt}{, }\citet{yue2023large}{, }P$^3$Defer\cite{zhang2024privacy}{, }Route{LLM} \cite{ong2024routellm}{, }SlimPM\cite{tan2024small}{, }Division-of-thoughts\cite{shao2025division}{, }R2R\cite{fu2025r2r}{, }CITER\cite{zheng2025citer}{, }MixLLM\cite{wang2025mixllm}{, }MoE$^2$\cite{jin2025moe}{, }Beagle\cite{zhong2025cross}{, }LookAhead Reasoning\cite{fu2025scaling}{, }APD\cite{israel2025accelerating}{, }LLMCAD\cite{xu2023llmcad}{, }Edge–cloud Hybrid\cite{hao2024hybrid}{, }PICE\cite{zhan2025pice}{, }FS-GEN\cite{zhang2024fast}{, }BestRoute\cite{ding2025bestroute}{, }AgentTTS\cite{wang2025agenttts},leaf,fill=myyellow]
      ]
    ]
    [Cloud-Edge Privacy\\ \S \ref{sec:collaboration_privacy}, edge=black!50, leaf3, fill=softblue
      [Inference Stage \\ \S \ref{sec:privacy_inference}, leaf1, fill=softblue,
        [Casper\cite{chong_casper_2024}{, }LLM Gatekeeper \cite{uzor_guarding_2025}{, }LLM-Anonymizer\cite{wiest_anonymizing_2024}{, }\citet{hartmann-etal-2024-llms}{, }CORE\cite{fan2025core}{, }P$^3$Defer\cite{zhang2024privacy}{, }RemoteRAG\cite{cheng_remoterag_2024}{, }CoGenesis\cite{zhang2024cogenesis}{, }PrivacyBoost-SLM\cite{zhang_enhancing_2024}{, }Apple Intelligence\cite{apple_foundation_models},leaf,fill=softblue]
      ]
      [Fine-tuning Stage \\ \S \ref{sec:privacy_train}, leaf1, fill=softblue,
        [MiniLLM\cite{gu2023minillm}{, }LlamaDuo\cite{park-etal-2025-llamaduo}{, }DRAG\cite{chen2025dragdistillingragslms}{, }HomeLLaMA\cite{homellama2025}{, }On-device Personalization\cite{qin2024enabling}{, }TinyLLM\cite{tianAnswers2024a}{, }Mix Distillation\cite{liSmall2025}{, }ADPA\cite{gaoAdvantageGuided2025}{, }CombLM\cite{ormazabalCombLMAdaptingBlackBox2023}{, }Logit Offsets\cite{liu2024tuning, prada2025}{, }CPT\cite{heCPTConsistentProxy2024}{, }GradOT\cite{yaoGradOTTrainingfreeGradientpreserving2025}{, }GOOD\cite{fangGOODDecodingTimeBlackBox2024}{, }PHLoRA\cite{vasaniPHLoRADatafreePosthoc2025}{, }LoRASuite\cite{liLoRASuiteEfficientLoRA2025}{, } LoRA-X\cite{farhadzadehLoRAXBridgingFoundation2025}{, }Trans-LoRA\cite{wangtextitTransLoRA2024}{, }Cross-LoRA\cite{xiaCrossLoRADataFreeLoRA2025}{, }CrossLM\cite{deng2025crosslm}{, }LSRP\cite{lsrp2025}{, }FedMKT\cite{fan-etal-2025-fedmkt}{, }FedPT\cite{gaoFedPTFederatedProxyTuning2024}{, }FedPFT\cite{pengFedPFT2024}{, }FDLoRA\cite{qiFDLoRAPersonalizedFederated2024}{, }FedPipe\cite{fangAutomatedFederatedPipeline2024}{, }FLoRA\cite{wangFLoRA2024},leaf,fill=softblue]
      ]
    ]
    [Trustworthiness \\ \S \ref{sec:collaboration_trust}, edge=black!50, leaf3, fill=softgray
      [Safety-Guided Decoding \\ \S \ref{sec:safety_guided_decoding}, leaf1, fill=softgray,
        [Purifying LLMs\cite{li2024purifying}{, }MOD\cite{shi2024decoding}{, }Logit offset\citet{mitchell2024an}{, }Offset Unlearning\cite{huang2025offset}{, }Weak-to-Strong Jailbreak\cite{zhao2025weaktostrong},leaf,fill=softgray]
      ]
      [SLM as Guardian \\ \S \ref{sec:guardian}, leaf1, fill=softgray,
        [Llama Guard\cite{inan2023llama, metallamaguard2, dubey2024llama, fedorov2024llama, chi2024llama, llama_guard_4}{, }Prompt Guard\cite{meta2024promptguard,llama-prompt-guard2}{, }ShieldGemma\cite{zeng2024shieldgemma, zeng2025shieldgemma}{, }WildGuard\cite{han2024wildguard}{, }ThinkGuard\cite{wen-etal-2025-thinkguard}{, }MiniCheck\cite{tang2024minicheck}{, }Guardrails\cite{nagireddy2024doubt}{, }LlamaFirewall\cite{chennabasappa2025llamafirewall},leaf,fill=softgray]
      ]
    ]
  ]
\end{forest}
\caption{A Taxonomy of SLM-LLM Collaboration.}
\label{fig:taxonomy}
\end{figure*}
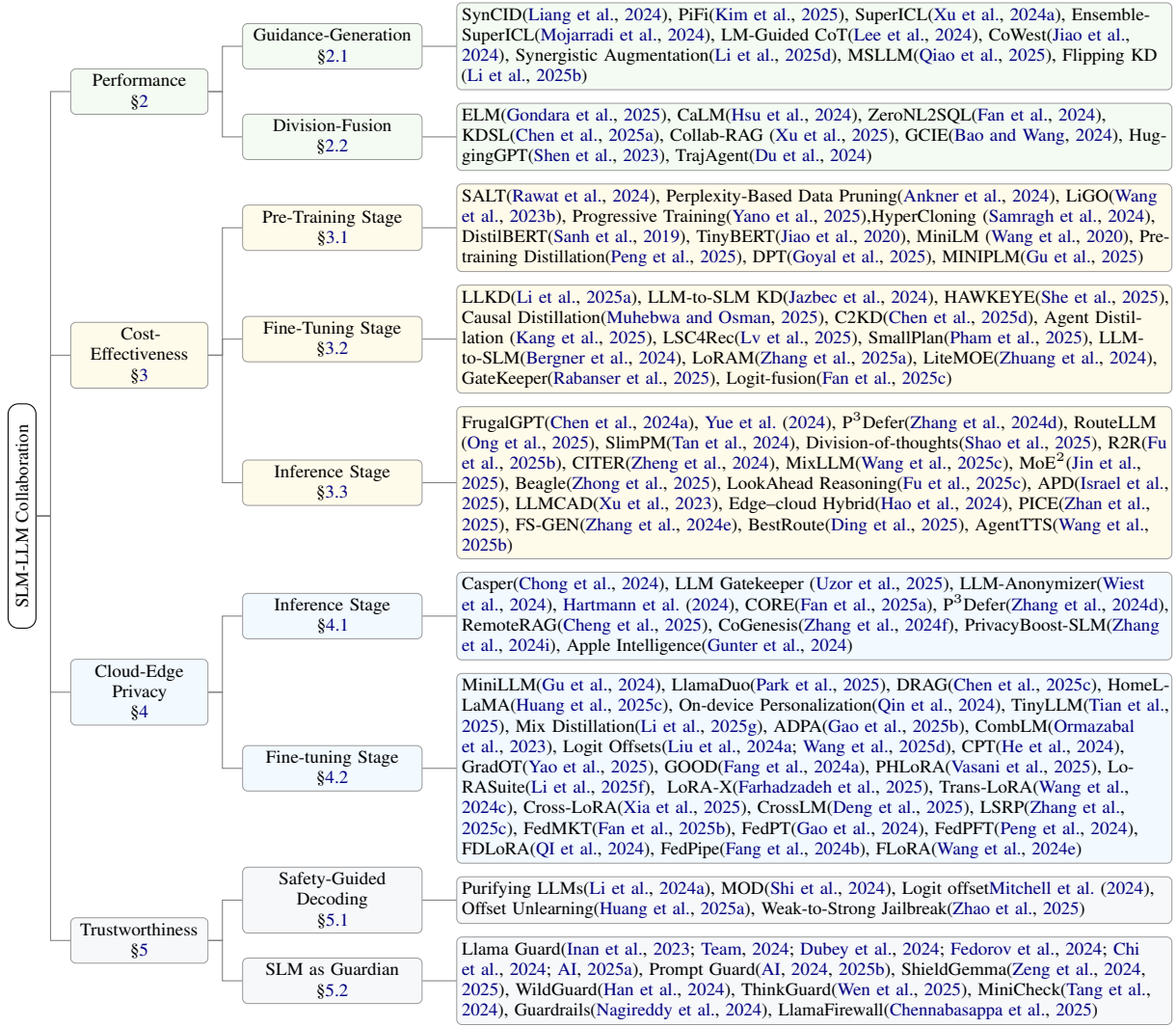

Researchers have developed various approaches to enable collaboration between SLMs and LLMs, leveraging the strengths of SLMs in customization, efficiency, and local deployment, alongside the powerful generalization capabilities of LLMs \cite{xu2023small, chen2024frugalgpt, wang2025agenttts, gu2023minillm, inan2023llama}. 
Despite the importance of this topic and the various approaches developed, there is a lack of a systematic survey centered on the \textbf{objectives of SLM--LLM collaboration}. Hence, we fill this gap by providing a taxonomy that unifies recent works under clear collaboration objectives.
Existing studies generally pursue four main objectives. First, by combining domain-specific SLMs that outperform general-purpose LLMs in specialized contexts, many works target collaboration for enhanced \textit{task performance} across both general and domain-specific tasks. Second, emphasizing \textit{cost-effectiveness}, SLMs are employed for lightweight processing, while LLMs are invoked selectively when necessary. Third, under the paradigm of \textit{cloud--edge collaboration}, on-device SLMs process privacy-sensitive data, whereas cloud-based LLMs provide general reasoning, ensuring both effectiveness and privacy. Finally, SLMs act as lightweight \textit{safety policy encoders}, while LLMs serve as backbone generators to improve output \textit{trustworthiness}.
Therefore, our survey is organized around the objectives of SLM--LLM collaboration: \textit{performance}, \textit{cost-effectiveness}, \textit{cloud--edge privacy}, and \textit{trustworthiness}. Under this framework, we propose a taxonomy of existing studies (in Fig.~\ref{fig:taxonomy} and Tab.~\ref{tab:slm_llm_overall}), summarize key insights, and outline future directions for efficient, secure, and deployable intelligent systems.

\paragraph{Differences from existing surveys}
Several surveys have examined SLMs in the LLM era.
\citet{wang2024comprehensive, wang2025survey} provided a broad overview of SLM design, acquisition, applications, and trustworthiness but mentioned SLM--LLM collaboration only briefly.
\citet{lu2024small} evaluated SLM advantages through experimental analysis, \citet{van2024survey} focused on architectures and compression techniques, and \citet{xu2024device} discussed on-device deployment, all with limited discussion of collaboration.
Existing collaboration surveys also remain narrow:
\citet{chen2024role} explored one-way collaboration (SLMs for LLMs or LLMs for SLMs), whereas \citet{niu2025collaborative, li2025collaborative} examined cloud--edge collaboration without addressing diverse collaboration objectives.
To fill this gap, we systematically review existing studies from the perspective of collaboration objectives, summarizing key insights to guide future SLM--LLM collaboration.

Our main contributions are summarized as follows: (1) We present the first systematic survey of SLM–LLM collaboration from a novel perspective centered on collaboration objectives, encompassing performance, cost-effectiveness, cloud–edge privacy, and trustworthiness. The overall taxonomy is illustrated in Fig. \ref{fig:taxonomy}.
(2) We analyze and categorize performance-oriented collaboration studies based on how they exploit the complementary strengths of SLMs and LLMs, forming two major paradigms: Guidance–Generation and Division–Fusion.
(3) We summarize cost-effectiveness-driven collaboration along the lifecycle of language models, covering the pre-training, fine-tuning, and inference stages.
(4) We review privacy-preserving collaboration between cloud and edge models according to how edge data are privately leveraged by cloud models during both the inference and fine-tuning stages.
(5) We discuss trustworthiness-oriented collaboration under two paradigms: safety-guided decoding and guardian–generator frameworks.

\begin{table*}[h!]
\centering
\tiny
\renewcommand{\arraystretch}{1.06}
\setlength{\tabcolsep}{3.5pt}
\caption{Representative works on collaborations across performance, cost-effectiveness, privacy, and trustworthiness.}
\label{tab:slm_llm_overall}
\begin{tabularx}{\linewidth}{l l l l l X}
\toprule
\textbf{Collab. Mechanism} & \textbf{Paper} & \textbf{SLM} & \textbf{LLM} & \textbf{Evaluation Dataset} & \textbf{Goal / Contribution} \\
\midrule
\multicolumn{6}{c}{\textbf{I. Collaboration for Performance}} \\
\midrule
\textbf{LLM guided SLM} & SynCID~\citeyearpar{liang2024synergizing} & BERT & GPT-3.5 & Banking77, CLINC150 & Refine labels for intent detection \\
\textbf{LLM guided SLM} & PiFi \citeyearpar{kim-etal-2025-plug}  & BERT, T5 & LLaMA~3.1-8B & SST-2, IMDB, \emph{etc.} & Frozen LLM layer enhances SLM \\
\textbf{SLM guided LLM} & SuperICL~\citeyearpar{xu2023small} & RoBERTa & GPT-3.5/4 & SST-2, MNLI, \emph{etc.} & Inject SLM predictions and confidence \\
\textbf{SLM guided LLM} & Ensem. SuperICL~\citeyearpar{mojarradi2024improving} & BERT/RoBERTa & GPT-3.5/LLaMA & SST-2, MNLI, \emph{etc.} & Multi-SLM guidance for LLMs \\
\textbf{SLM guided LLM} & LM-Guided CoT~\citeyearpar{lee2024can} & Flan-T5 & ChatGPT & HotpotQA, 2WikiMultiHopQA & Provide rationales to improve reasoning \\
\textbf{SLM guided LLM} & CoWest~\cite{jiao2024synergistic} & LLaMA-3-8B & GPT-4/3.5 & IfQA, MedMCQA, \emph{etc.} & Preference-aligned weak-to-strong collab. \\ 
\textbf{Parallel} & ELM~\citeyearpar{gondara2025elm} & ClinicalBERT & GPT-4 & BCCR Pathology Reports & Aggregate SLM votes; LLM verifies \\
\textbf{Parallel} & CaLM~\citeyearpar{hsu2024calm} & T5, Flan-T5 & Flan-PaLM, GPT-3.5 & QAGS-CNNDM, Q\textsuperscript{2}, \emph{etc.} & Cross-verification SLM$\leftrightarrow$LLM \\
\textbf{Sequential} & ZeroNL2SQL~\citeyearpar{fan2024combining} & T5, BART & GPT-3.5/4 & Spider, WikiSQL, \emph{etc.} & SLM drafts SQL; LLM finalizes \\
\textbf{Sequential} & KDSL~\citeyearpar{chen2025knowledge} & Qwen2-VL 2B & Qwen2-VL 7B & -- & SLM rules; LLM inference \\
\textbf{Sequential} & HuggingGPT~\citeyearpar{shen2023hugginggpt} & CLIP, BLIP, T5 & GPT-3.5 & ImageNet, COCO, \emph{etc.} & LLM orchestrates multimodal SLMs \\
\midrule
\multicolumn{6}{c}{\textbf{II. Collaboration for Cost-effectiveness}} \\
\midrule
\textbf{Data curation} & SALT~\citeyearpar{rawat2024little} & Transformer-1.5B & Transformer-2.8B & Pile & SLM-guided data selection \\
\textbf{Data curation} & \citet{ankner2024perplexed} & MPT-125M & MPT-1B/3B & The Pile, Dolma, \emph{etc.} &  SLM perplexity-guided efficient pruning \\

\textbf{Co-curriculum} & \citet{yano2025efficient} & LLaMA-1B & LLaMA-8B & FineWeb-Ed & Joint curriculum design \\
\textbf{Co-curriculum} & LiGO \citeyearpar{wang2023learning} & BERT-S, DeiT-S & BERT-B, DeiT-B & GLUE, SQuAD, ImageNet & Learned growth reduces training cost \\
\textbf{Co-curriculum} & \textsc{HyperCloning}~\citeyearpar{samragh2024scaling} & OPT-350M & OPT-1.3B & DOLMA, The Pile, \emph{etc.} & Function-preserving SLM-to-LLM initial. \\

\textbf{Pre-train distill.} & \citet{peng2024pre} & GLM-4-1.9B & GLM-4-9B & MMLU, C-Eval, \emph{etc.} & Progressive knowledge transfer \\
\textbf{Pre-train distill.} & DPT~\citeyearpar{goyal2025distilled} & -- & -- & Synthetic bigram data & Distillation improves scaling, harms ICL \\
\textbf{Pre-train distill.} & \textsc{MiniPLM}~\citeyearpar{gu2024miniplm} & Qwen 200M/500M & Qwen1.5-1.8B & DCLM subset & Offline sampling accelerates pretraining \\

\textbf{Selective KD} & SynCID~\citeyearpar{liang2024synergizing} & BERT & text-davinci-003 & BANKING, CLINC & Latent-space alignment distillation \\
\textbf{Selective KD} &LLM-to-SLM~\citeyearpar{bergner2024think} & T5-Small & T5-Large & WMT, CNN, \emph{etc.} & Frozen LLM conditions fast SLM \\

\textbf{Proxy transfer} & LiteMoE~\citeyearpar{zhuang2024litemoe} & Gemma-2B & Gemma-7B & MMLU, GSM8K, \emph{etc.} & Efficient proxy tuning \\
\textbf{Proxy transfer} & GateKeeper~\citeyearpar{rabanser2025gatekeeper} & Gemma-2B & Gemma-7B & CIFAR, ARC, VQAv2, \emph{etc.} & Conf-tuned SLM improves cascading \\
\textbf{Proxy transfer} & G-Boost~\citeyearpar{fan2025g} & TinyLlama-1B & LLaMA2-13B & GSM8K, MATH-500 & Reward-guided dynamic collaboration \\

\textbf{Cascade Routing} & R2R~\citeyearpar{fu2025r2r} & DeepSeek-R1-1.5B & DeepSeek-R1-32B & AIME, GPQA & Token-level routing for efficiency \\
\textbf{Cascade Routing} &SlimPM~\citeyearpar{tan2024small} & LLaMA2-7B & LLaMA2-70B & NQ, TriviaQA, \emph{etc.} &  Proxy-guided selective retrieval efficiency \\
\textbf{Cascade Routing} & CITER~\citeyearpar{zheng2025citer} & Qwen2-1.5B & Qwen2-72B & CSQA, ARC, \emph{etc.} & Token-level routing boosts efficiency \\
\textbf{Cascade Routing} & MixLLM~\citeyearpar{wang2025mixllm}& Llama-3-8B & GPT-3.5/4, etc. & RouterBench & Adaptive routing \\

\textbf{Speculative} & Beagle~\citeyearpar{zhong2025cross} & LLaMA-68M & LLaMA-7B & ShareGPT & Cross-attention drafting \\
\textbf{Speculative} & APD~\citeyearpar{israel2025accelerating} & Qwen2.5-0.5B & Dream-7B  & GSM8K, GPQA, MATH & Adaptive parallel decoding via PoE \\
\textbf{Speculative} & LLMCAD~\citeyearpar{xu2023llmcad}& GPT-2/mT5 & Vicuna/LLaMA2 & CNN/DailyMail, \emph{etc.} & Token-tree drafting \\
\textbf{Speculative} & Cloud-edge~\citeyearpar{hao2024hybrid}& TinyLlama & GPT-3.5 & GSM8K & Token-level edge–cloud collaboration\\


\textbf{Optimal TTS} & AgentTTS~\citeyearpar{wang2025agenttts} & LLaMA-3-3B & LLaMA-3-70B & HotpotQA, CWQ, \emph{etc.} & Budget-aware collaboration \\
\textbf{Optimal TTS} & BestRoute~\citeyearpar{ding2025bestroute}  & Phi-3-mini/medium, \emph{etc.} & GPT-4o/3.5 & MixInstruct, RewardBench, \emph{etc.} & Adaptive routing minimizes cost \\

\midrule

\multicolumn{6}{c}{\textbf{III. Privacy-aware Cloud–Edge Collaboration}} \\
\midrule
\textbf{PII Gatekeeper} & \citet{uzor_guarding_2025} & Phi-3.5, Gemma-2 & GPT-4o & HealthTap, MeQSum & Rewrite prompts to remove PII \\
\textbf{PII Gatekeeper}& LLM-Anonymizer \citeyearpar{wiest_anonymizing_2024} & Llama-3 8B, \emph{etc.} & Any & real clinical letters & Local LLM anonymization \\
\textbf{PII Gatekeeper} & P$^3$Defer \citeyearpar{zhang2024privacy} & Gemma-2B & Gemma-7B & GSM8K, MedSum, \emph{etc.} & CoT policy RL reduces leakage \\


\textbf{All-info Guardian} & CoGenesis~\citeyearpar{zhang2024cogenesis} & StableLM-Zephyr & GPT-4-turbo & Avocado Research Email & Cloud knowledge for edge inference \\
\textbf{All-info Guardian} & \citet{hartmann-etal-2024-llms} & Gemini 1.0 Nano-2 & Gemini 1.0 Ultra & GSM8K & Cloud ICL demos for edge SLM \\
\textbf{All-info Guardian} & PrivacyBoost-SLM \citeyearpar{zhang_enhancing_2024} & BioLinkBERT, \emph{etc.} & GPT-3.5 & MedQA, HEADQA, \emph{etc.} & Privacy prompting boosts medical QA
 \\

\textbf{SLM as Learner} & MiniLLM~\citeyearpar{gu2023minillm} & GPT-2-125M & GPT-2-1.5B & DollyEval, SelfInst, \emph{etc.} & Distill LLM into SLM \\
\textbf{SLM as Learner} & Mix Distillation~\citeyearpar{liSmall2025} & Qwen2.5-0.5B, \emph{etc.} & QwQ-32B, \emph{etc.} & MATH, GSM8K, \emph{etc.} & Learnability gap, mixed distillation \\

\textbf{LLM as Learner} & Proxy-tuning~\citeyearpar{liu2024tuning} & LLaMA-2-7B & LLaMA-2-70B & GSM, Toxigen, \emph{etc.} & Steer LLM with private SLM \\

\textbf{LLM as Learner} & Cross-LoRA~\citeyearpar{xiaCrossLoRADataFreeLoRA2025} & Qwen2.5-1.5B/3B, \emph{etc.} & Hetero-families & ARC-c, ARC-e, OBQA, \emph{etc.} & Zero-data subspace LoRA transfer \\

\textbf{Co-learners} & FDLoRA~\citeyearpar{qiFDLoRAPersonalizedFederated2024} & LLaMA-2-7B & LLaMA-2-7B & BGL, Spirit, \emph{etc.} & Federated LoRA with sharing \\
\textbf{Co-learners} & FedMKT~\citeyearpar{fan-etal-2025-fedmkt}& GPT-2-xl, OPT, \emph{etc.} & LLaMA2-7B & RTE, WIC, BoolQ, \emph{etc.} & Bidirectional federated co-distillation\\

\midrule

\multicolumn{6}{c}{\textbf{IV. Collaboration for Trustworthiness}} \\
\midrule
\textbf{Safe decoding} & MOD~\citeyearpar{shi2024decoding} & Safety branch & LLaMA-2-13B/70B & -- & Logit fusion for safety–utility \\
\textbf{Safe decoding} & W2S Jailbreaking~\citeyearpar{zhao2025weaktostrong} & LLaMA-2-13B & LLaMA-2-70B & AdvBench, MaliciousInstruct & Analyze SLM logit effects \\
\textbf{Safe decoding} & Offset Unlearning~\citeyearpar{huang2025offset} & LLaMA-2-7B  & LLaMA-2-13B & TOFU & Logit offset for unlearning \\
\textbf{Safe decoding} & LLM Purifying~\citeyearpar{li2024purifying} & CodeLLaMA-7B & CodeLLaMA-13B & HumanEval, LAMBADA, \emph{etc.} & Logit fusion for clean gen. \\
\textbf{SLM guard} & LLaMA Guard~\citeyearpar{inan2023llama} & LLaMA-2-7B & LLaMA-70B & ToxicChat, OpenAI ModEval & I/O safety screening \\
\textbf{SLM guard} & Prompt Guard / 2~\citeyearpar{meta2024promptguard, llama-prompt-guard2} & DeBERTa-v2-22M/86M & LLaMA/GPT-family & Red-teaming (multilingual), \emph{etc.} & Lightweight input filtering \\
\textbf{SLM guard} & ShieldGemma / 2~\citeyearpar{zeng2024shieldgemma, zeng2025shieldgemma} & Gemma-2B/9B/27B & Any & ToxicChat, OpenAI Mod, \emph{etc.} & Fine-grained moderation \\
\textbf{SLM guard} & WildGuard~\citeyearpar{han2024wildguard} & Mistral-7B-v0.3 & Any & WildGuardMix & Multi-risk detection \\
\textbf{SLM guard} & ThinkGuard~\citeyearpar{wen-etal-2025-thinkguard} & LLaMA-3-8B & Any & BeaverTails, ToxicChat & Reasoned moderation \\
\textbf{SLM guard} & MiniCheck~\citeyearpar{tang2024minicheck} & Flan-T5 / UL2 & GPT-4 / Claude & LLM-AggreFact & Fact verification \\
\bottomrule
\end{tabularx}
\end{table*}



\section{Collaboration for Performance}
\label{sec:collaboration_performance}

The optimal LLM varies across tasks and queries \cite{jiang2023llm, ashiga2025ensemble, bell-2025-less, wang2025agenttts}, making single-model solutions suboptimal. Leveraging complementary models is therefore effective. Owing to their customizability, SLMs have produced many domain-specific products, such as SciGLM \cite{zhang2024sciglm}, ChemLLM \cite{zhang2024chemllm}, and Biomistral \cite{labrak2024biomistral}. Hence, SLM–LLM collaboration for performance has been a prevalent strategy. We divide existing works mainly into two paradigms: (1) \textit{Guidance–Generation}, where one model steers another’s generation, and (2) \textit{Division–Fusion}, where the SLMs and LLMs leverage their complementary strengths on distinct tasks.



\subsection{Guidance–Generation Collaboration}
\label{sec:advisory}
Guidance--Generation collaboration involves one assistant model providing guidance based on its strengths to support a backbone model in generation. When one model performs well overall and another excels in a specific aspect, the former serves as the generator while the latter offers complementary guidance. This paradigm generally appears in two forms: (1) LLM-guided SLM generation and (2) SLM-guided LLM generation.

\begin{figure}[t]
  \centering
  \includegraphics[width=0.99\linewidth]{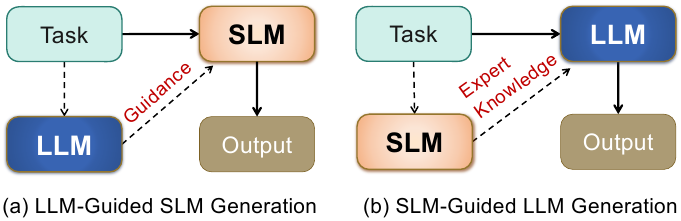}  
  \caption{Guidance-Generation Collaboration.}
  \label{fig:guidance}
\end{figure}

\paragraph{(1) LLM-guided SLM generation.} 
As shown in Fig.~\ref{fig:guidance}(a), the LLM uses its semantic understanding to provide fine-grained guidance for task-specific SLMs. For example, SynCID~\cite{liang2024synergizing} employs LLM-generated task descriptions to guide SLM reasoning. PiFi \cite{kim-etal-2025-plug} uses LLMs' certain knowledge layer plug-in to enhance SLM’s generalization and understanding.

\paragraph{(2) SLM-guided LLM generation.} 
As shown in Fig.~\ref{fig:guidance}(b), the SLM offers domain expertise and cues, while the LLM integrates this information for more accurate and reliable outputs. For instance, SuperICL and its variant~\cite{xu2023small, mojarradi2024improving} inject SLM predictions and confidence scores into the LLM’s context, and LM-Guided CoT~\cite{lee2024can} uses SLM-generated reasoning chains to guide LLM inference. 
CoWest~\cite{jiao2024synergistic} aligns weak–strong preferences to synergize domain expert SLMs' knowledge and LLMs' general reasoning. Synergistic Augmentation \cite{li-etal-2025-synergistic} uses SLM to guide LLM with confident demonstrations and slot hints, boosting zero-shot slot filling performance. 
MSLLM \cite{qiao-etal-2025-mixture} combines SLM’s precise error detection with LLM’s fluency and semantic understanding for superior Chinese spelling correction.
Flipping KD \cite{liFlippingKnowledgeDistillation2025} enables LLMs to learn task-specific ability from SLMs. 

\subsection{Division–Fusion Collaboration}
\label{sec:cospecialized_collaboration}
Division--fusion collaboration paradigm is preferred when multiple models exhibit heterogeneous capabilities and no single backbone model is highly suitable. According to different task needs, this paradigm mainly includes (1) \textit{Parallel Ensemble} for collective intelligence and (2) \textit{Sequential Cooperation} for multi-stage workflows.



\begin{figure}[t]
  \centering
  \includegraphics[width=0.99\linewidth]{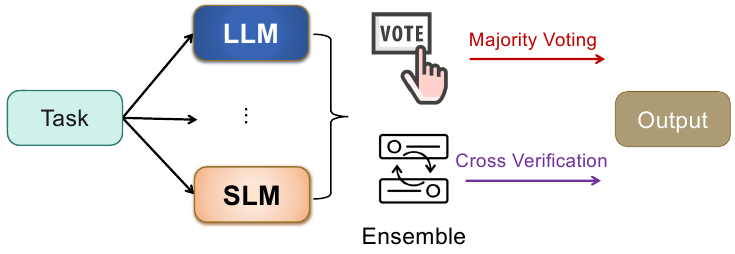}  
  \caption{Parallel Ensemble: Collective Intelligence.}
    \vskip -1em
  \label{fig:ensemble}
\end{figure}

\paragraph{(1) Parallel Ensemble: Collective Intelligence}
As shown in Fig.~\ref{fig:ensemble}, multiple SLMs and LLMs work in parallel, and their outputs are integrated for higher accuracy. 
Two common strategies are \textit{majority voting} and \textit{cross-verification}.
For \textit{majority voting}, as in ELM~\cite{gondara2025elm}, multiple independent solutions are sampled from a set of distinct models and combined by consensus to integrate multi-model capabilities.
For \textit{cross-verification}, as in CaLM~\cite{hsu2024calm}, an LLM and an SLM independently generate answers in parallel and iteratively refine both inputs and model states, continuing the process until their outputs reach consensus.

\begin{figure}[b]
  \centering
  \includegraphics[width=0.99\linewidth]{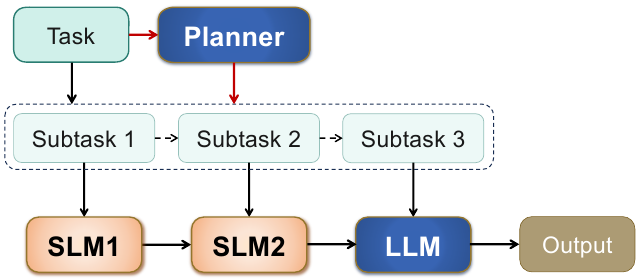}  
  \caption{Sequential Cooperation: Division of Labor.} 
  \label{fig:division}
\end{figure}

\paragraph{(2) Sequential Cooperation: Division of Labor}
For multi-stage tasks, subtasks are assigned to suitable models and connected in a pipeline, as shown in Fig.~\ref{fig:division}. When tasks are explicitly staged, SLMs and LLMs take different roles: SLMs handle precise components (e.g., schema matching in ZeroNL2SQL~\cite{fan2024combining} and KDSL~\cite{chen2025knowledge}, question decomposition in Collab-RAG \cite{xu2025collab}), while LLMs manage complex reasoning (e.g., GCIE~\cite{bao2024general}). 
When stages are implicit, the LLM acts as a \textit{planner} and SLMs as \textit{executors}, as in HuggingGPT~\cite{shen2023hugginggpt} and TrajAgent~\cite{du2024trajagent}, which assign subtasks to expert SLMs and integrate them. 

\subsection{Discussion}
Current SLM–LLM collaborations for performance highlight the potential of combining specialized and general models but remain limited by fragmented ecosystems and incomplete evaluation standards. Future research should establish open, cross-domain platforms enabling efficient model discovery and integration, alongside collaborative benchmarks that assess structured cooperation, throughput, and cost–performance trade-offs to better capture real-world system efficiency.

\section{Collaboration for Cost-Effectiveness Trade-off}
\label{sec:collaboration_cost_effectiveness}




LLMs deliver advanced performance but face prohibitive costs, including computational expense (FLOPs, GPU hours), high latency, significant communication overhead, large storage footprints, and substantial API spend. Mitigating these costs through isolated model optimization is often inefficient and reaches diminishing returns. SLMs offer a compelling pathway to cost-efficiency by serving as lightweight, specialized collaborators~\cite{wang2025survey}. This section targets \emph{cost-effectiveness} in SLM-LLM collaboration frameworks, which are organized across the language modeling lifecycle into three stages: (1) 
\emph{Pre-training} Stage: strategies like teacher-guided data curation and co-curricula use LLMs to guide SLM learning for improved compute and data efficiency;
(2) \emph{Tuning} stage: selective knowledge distillation (LLM$\!\rightarrow$SLM) and proxy transfer (SLM$\!\rightarrow$LLM warm-starts or adapters) cut adaptation FLOPs and storage while preserving task quality; 
and (3) \emph{Inference} stage, cascade routing, speculative decoding, and compute-optimal test-time scaling allocate just-enough capacity per query to lower latency, communication, and per-call/API cost. 


\subsection{Collaboration During Pre-Training Stage}
\label{sec:cost_effectiveness_pretrain}

\begin{figure}[t]
    \centering
    \begin{subfigure}[t]{\linewidth}
        \centering
 \includegraphics[width=0.9\linewidth]{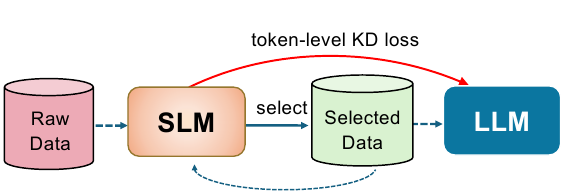}
        \caption{SLM selects/reshapes data and feeds it into LLM}
        \label{fig:dc1}
    \end{subfigure}
    \begin{subfigure}[t]{\linewidth}
        \centering
    \includegraphics[width=0.9\linewidth]{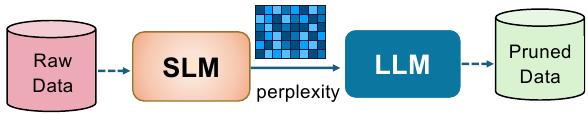}
        \caption{Perplexity-based data pruning with SLM}
        \label{fig:dc2}
    \end{subfigure}%
    \caption{Data Curation.}
    \label{fig:dc}
\end{figure}

\begin{figure}[b]
    \centering
    \begin{subfigure}[t]{0.44\linewidth}
        \centering
        \includegraphics[width=\linewidth]{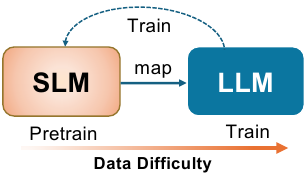}
        \caption{Co-Curriculum}
        \label{fig:co-curriculum}
    \end{subfigure}
    \hfill
    \begin{subfigure}[t]{0.48\linewidth}
        \centering
        \includegraphics[width=\linewidth]{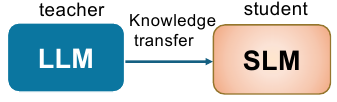}
        \caption{Pre-training Distillation}
        \label{fig:kd}
    \end{subfigure}
    \caption{(a) Co-curriculum learning; (b) Knowledge distillation.}
    \label{fig:pretrain_methods}
\end{figure}

The pre-training stage constitutes a computationally intensive phase of the LLM lifecycle, making it a primary target for cost-effectiveness optimization. The challenge lies in achieving high model capability without incurring the prohibitive expense of training on a massive yet low-quality data corpus. SLM–LLM collaboration during the pre-training stage addresses this by employing inter-model guidance to improve data efficiency and optimize the training trajectory. These methods can be categorized into three types: (1) Data curation, where Fig.~\ref{fig:dc1} illustrates an LLM refining data for efficient SLM pre-training, and Fig.~\ref{fig:dc2} shows an SLM evaluating data quality to reduce LLM pre-training cost; (2) Co-curriculum (Fig.~\ref{fig:co-curriculum}), which progressively grows from small to large model capacity; and (3) Pre-training distillation (Fig.~\ref{fig:kd}), which directly transfers knowledge from a teacher large model to a student small model.
\paragraph{(1) Data Curation}
High-quality data are vital for language model training, yet real-world corpora are often low-quality and noisy. Using low-quality data inflates computational costs and degrades model performance. SLM–LLM collaboration enhances data curation cost-effectively: (1) SLMs select or reshape data for LLM pretraining to boost performance with less data (Fig.~\ref{fig:dc1}); (2) SLMs estimate data quality (e.g., perplexity) to guide LLM pretraining efficiently (Fig.~\ref{fig:dc2}).
For instance, the SALT~\cite{rawat2024little} shows that a compact model can supply soft labels and identify informative hard examples, accelerating large-model pre-training while improving quality, reducing steps, FLOPs, and energy; Conversely, \citet{ankner2024perplexed} adopts perplexity-based filtering with an SLM as the reference scorer to remove low-yield documents/tokens, cutting the optimization steps needed to reach a target loss. 

\paragraph{(2) Co-Curriculum}
Training model families efficiently is challenging because large models are costly to start from scratch. Co-curriculum mitigates this through this agenda: pretrain a small model, grow it to a larger one via parameter reuse, then continue training on equal-or-harder data; repeat the grow $\to$ train cycle to form a consistent model family across sizes (see Fig.~\ref{fig:co-curriculum}). 
Growth can be realized in several ways: \textsc{LiGO~\cite{wang2023learning}} learns a linear growth operator that maps small-model weights to a larger architecture, where the operator is factorized into width/depth with Kronecker structure before continuing standard training, yielding up to ~50\% compute savings compared to large-from-scratch while preserving quality. 
\textsc{Progressive training}~\cite{yano2025efficient} expands width/depth in stages (e.g., 1B$\to$2B$\to$8B), carrying weights forward and using size-aware schedules that assign higher max LR for smaller models, longer warmup for larger ones, and token allocation that favors stages with higher marginal gain. 
\textsc{HyperCloning} \cite{samragh2024scaling} performs a hidden-dimension expansion preserving function by cloning linear layers so that the larger model initially reproduces the smaller model’s logs, producing stable warm starts and substantial savings in GPU-hours compared to cold starts.

\paragraph{(3) Pre-Training Distillation}
Pre-training distillation (PD) transfers knowledge from a teacher LLM to a student SLM through logit matching and data refinement, reducing data requirements while enhancing model capability  (see Fig.~\ref{fig:kd}).
Existing works fall into two categories based on the supervision signal: (i) logits-based and (ii) data-based pre-training distillation.
For the former, task-agnostic distillation frameworks such as \textsc{DistilBERT}~\cite{Sanh2019DistilBERT}, \textsc{TinyBERT}~\cite{jiao2019tinybert}, and \textsc{MiniLM}~\cite{Wang2020MiniLM} transmit teacher logits and embeddings to students as supervision signals.
\citet{peng2024pre} extend this to larger student LMs, addressing challenges in logits storage, compression, signal design, capacity gap, and online–offline trade-offs. 
\textsc{DPT}~\cite{goyal2025distilled} further shows logits-based pre-training distillation enhances test-time scaling but weakens in-context learning, providing analytical insights into this trade-off.
For the latter, \textsc{MiniPLM}~\cite{gu2024miniplm} performs offline teacher inference to reshape the pre-training corpus by (a) difficulty reweighting, prioritizing examples with informative teacher signals, and (b) diversity balancing, expanding topical, stylistic, and structural coverage, then trains a compact student on a refined mix.


\subsection{Collaboration During Tuning Stage}
\label{sec:cost_effectiveness_tuning}
\begin{figure}[t]
    \centering
\includegraphics[width=0.6\linewidth]{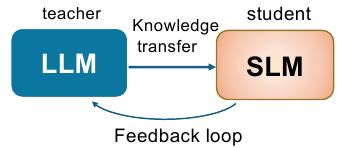}
    \caption{Selective Knowledge Distillation. 
    }
    \label{fig:tune1}
\end{figure}

\begin{figure}[b]
    \centering
    \begin{subfigure}[t]{\linewidth}
        \centering
 \includegraphics[width=0.8\linewidth]{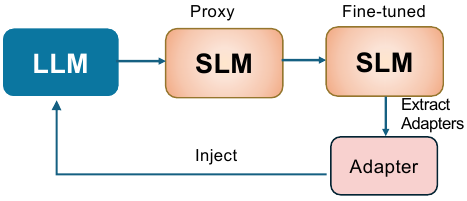}
        \caption{SLM-trained Compatible Parameters}
        \label{fig:proxy1}
    \end{subfigure}
    \begin{subfigure}[t]{\linewidth}
        \centering
    \includegraphics[width=0.95\linewidth]{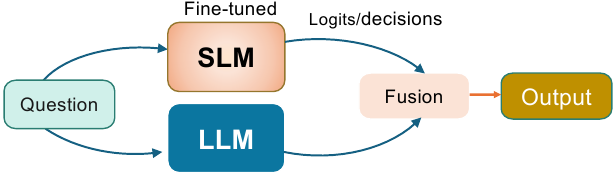}
        \caption{SLM-output Merging}
        \label{fig:proxy2}
    \end{subfigure}%
    \caption{Proxy Transfer.}
    \label{fig:proxy}
\end{figure}

Fine-tuning language models is costly along two fronts: (1) for tuning SLMs, relying on LLM distillation signals during adaptation increases FLOPs or API expenses, and fine-tuning quality depends on supervision quality: poor data wastes both tokens and computation
and (2) directly fine-tuning the LLM is expensive in terms of FLOPs, memory, and time. SLM–LLM collaboration mitigates these costs by: (1) \emph{Selective Knowledge Distillation} (SLM~$\Leftarrow$~LLM) replaces repeated LLM usage with distilled SLMs and increases information per token via selective signals and sampling (Fig.~\ref{fig:tune1}); and (2) \emph{Proxy Transfer} (SLM~$\Rightarrow$~LLM) employs SLMs to produce reusable updates or control signals for efficient LLM adaptation (Figs.~\ref{fig:proxy1} and~\ref{fig:proxy2}).

\paragraph{(1) Selective Knowledge Distillation.}
Full teacher supervision is costly and often redundant, while selective KD focuses limited guidance where it moves the needle most. We frame selective KD as teacher-in-the-loop curation that minimizes adaptation cost while preserving quality (Fig.~\ref{fig:tune1}). The pipeline selects high-yield samples where the teacher is confident but the student is uncertain, shapes supervision beyond hard labels to raise information per token, targets imitation to task-relevant representations instead of uniform mimicry, and then closes the loop using student feedback to guide the next round of teacher queries. In this loop, adaptive selection steers supervision to the error frontier—reducing data and teacher calls (LLKD~\cite{li2024learning}; student-in-the-loop selection with retained rationales~\cite{jazbec2024efficient}); dense yet compact supervision replaces bare labels with short, information-rich signals that clarify boundaries and reasoning—contrastive label refinement (SynCID~\cite{liang2024synergizing}), concise CoT (HAWKEYE~\cite{she2025hawkeye}, and causal explanations for robustness~\cite{muhebwa2025causal, lore2024large}); targeted matching aligns losses with where features live—layers/heads and cross-head logits that carry discriminative structure (C2KD~\cite{chen2025c2kd}); and behavioral KD distills compact tool-use trajectories to amortize exploration (Agent Distillation~\cite{kang2025distilling}). Applications instantiate the same loop end-to-end: LLM-assisted candidate generation with on-device ranking (LSC4Rec~\cite{lv2025collaboration}) operationalizes adaptive selection in recommendation systems; latency-sensitive recommenders pair KD with pruning/quantization~\cite{behdin2025efficient} to shape supervision under strict budgets; lightweight planners distilled via LLM-guided SFT/RL (SmallPlan~\cite{pham2025smallplan}) emphasize targeted matching to task-relevant skills; and “single frozen-LLM prompt, SLM decode” amortization~\cite{bergner2024think} closes the loop by throttling future teacher calls based on student performance—collectively cutting FLOPs, tokens, and LLM calls while retaining near-LLM quality.

\paragraph{(2) Proxy Transfer}
Proxy transfer reduces LLM adaptation costs by using small models as proxies to generate parameters or signals that can be reused by the larger model. This strategy operates through two main pathways: (1) SLMs, often compressed from the LLM, are fine-tuned with low resources to produce LoRA-like updates compatible with the original LLM (Fig.~\ref{fig:proxy1}); and (2) fine-tuned SLMs have their outputs merged with those of the LLM, simulating the effect of fine-tuning the large model through a lightweight proxy (Fig.~\ref{fig:proxy2}). Both pathways reduce the LLM’s required FLOPs, data, and tuning time.
For example, LoRAM~\cite{zhang2025train} trains low-rank adapters on an SLM to locate effective subspaces, later reused in LLMs. LiteMOE~\cite{zhuang2024litemoe} extracts proxy submodels for on-device tuning, enabling local specialization and feedback propagation. GateKeeper~\cite{rabanser2025gatekeeper} cascades an SLM and LLM to handle easy cases locally and defer harder ones, cutting inference cost, while G-Boost~\cite{fan2025g} integrates fine-tuned SLM outputs to refine LLM distributions with minimal updates.


\subsection{Collaboration During Inference Stage}
\label{sec:cost_effectiveness_inference}
\begin{figure}[b]
    \centering
\includegraphics[width=0.99\linewidth]{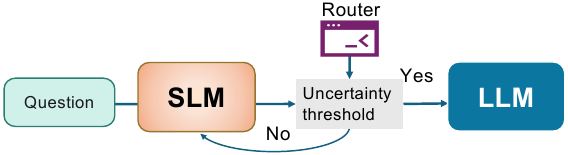}
    \caption{Cascade Routing.}
    \label{fig:inf1}
\end{figure}

\begin{figure}[t]
    \centering
\includegraphics[width=0.99\linewidth]{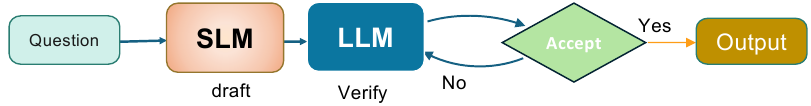}
    \caption{Speculative Decoding.}
    \label{fig:inf3}
\end{figure}

\begin{figure}[b]
    \centering
\includegraphics[width=0.99\linewidth]{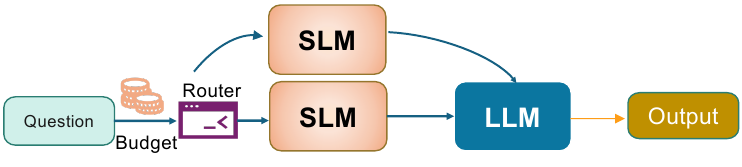}
    \caption{Compute-optimal TTS.}
    \label{fig:tts}
\end{figure}

Inference cost arises from three sources: (i) running large models on \emph{every} request (high FLOPs, latency, or API cost); (ii) decoding inefficiency (token-by-token verification by a single heavy model); and (iii) misallocated compute (overspending on easy queries and underspending on hard ones). SLM–LLM collaboration at the inference stage can address these through:
(1) Cascade routing: a lightweight router which is guided using some uncertainty threshold, sending most traffic to a cheap SLM and escalating only when necessary (Fig.~\ref{fig:inf1});
(2) Speculative decoding: a fast drafter (often an SLM) proposes multiple tokens that a strong verifier (LLM) accepts or corrects in bulk (Fig.~\ref{fig:inf3}); and
(3) Compute-optimal test-time scaling (TTS):
under a fixed budget, optimizes model choice and sampling budgets across SLM/LLM families to meet the target quality (Fig.~\ref{fig:tts}).

\paragraph{(1) Cascade Routing}
Cascade routing reduces average serving costs by directing most requests to a cheap SLM and escalating only complex cases to an LLM. Routers operate at different granularities: (i) query-level for holistic request classification, (ii) token/span-level for fine-grained generation control, and (iii) resource-aware for expert selection under system constraints.

\noindent
\textbfit{(i) Query-level routers} make a single deferral decision per input. Classical methods train a lightweight classifier to judge response reliability~\cite{chen2024frugalgpt,ong2024routellm,zhang2024privacy}. Extensions predict query difficulty for adaptive routing~\cite{yue2023large,hao2024hybrid}, including SlimPM~\cite{tan2024small}, which uses a slim proxy to detect an LLM’s knowledge gaps and decide when to retrieve.

\noindent\textbfit{(ii) Token/span-level routers} exert finer control by only invoking the LLM for critical tokens or segments. This improves the cost–quality trade-off by letting SLMs handle non-critical generation and LLMs only for critical tokens (e.g., CITER~\cite{zheng2025citer} and Roads to Rome \cite{fu2025r2r}).

\noindent\textbfit{(iii) Resource-aware routers} frame model selection as a constrained optimization problem, which uses contextual bandits to pick the best model under latency constraints~\cite{wang2025mixllm} or use MoE gates to dispatch requests across edge/cloud models, balancing load and budget~\cite{jin2025moe}.

\paragraph{(2) Speculative Decoding}
Speculative decoding reduces LLM compute by allowing a cheaper drafter to propose tokens (or short continuations) that a stronger verifier accepts in bulk (Fig.~\ref{fig:inf3}). Existing studies can be grouped into three major directions: (i) adaptive draft–verify pipelines, (ii) system- and hardware-aware parallelization, and (iii) distributed and hierarchical multi-model collaboration.

\noindent\textbfit{(i) Adaptive Draft–Verify Pipelines}
Early frameworks such as speculative decoding~\cite{leviathan2023fast,chen2023accelerating} and SpecDec~\cite{xia2023speculative} establish the draft–then–verify paradigm, where SLMs propose multi-token drafts and LLMs verify them in parallel for speedups. BiLD~\cite{kim2023speculative}, staged SD~\cite{spector2023accelerating}, and LLMCad~\cite{xu2023llmcad} extend this idea to big-little or on-device settings with rollback and correction strategies. Later methods including SpecInfer~\cite{miao2024specinfer}, cascade SD~\cite{chen2024cascade}, and SpecTr~\cite{sun2023spectr}, introduce tree- or cascade-based verification, while online SD~\cite{liu2024online}, TurboSpec~\cite{liu2024optimizing}, SCD~\cite{yuan2024speculative}, and mixture-of-attentions SD~\cite{zimmer2024mixture} add adaptive retraining, feedback control, and enhanced drafting accuracy. Together, these approaches evolve SD into dynamic, feedback-driven pipelines that balance quality and efficiency.

\noindent\textbfit{(ii) System- and Hardware-Aware Parallelization}
Several systems accelerate SD through pipeline and hardware optimization. EdgeLLM~\cite{xu2024edgellm}, SpecExec~\cite{svirschevski2024specexec}, and SPIN~\cite{chen2025spin} boost throughput via multi-token speculation, GPU pipelining, and heterogeneous SLM selection, while Dovetail~\cite{zhang2024dovetail} and APD~\cite{israel2025accelerating} exploit heterogeneous devices and adaptive speculative width. Hardware-level efforts like HADES~\cite{yang2025hades} embed SD primitives into accelerators, and HiSpec~\cite{kumar2025hispec} uses early-exit layers and cache reuse for lightweight intermediate verification. These methods make SD a pipeline-optimized, hardware-co-designed inference paradigm.

\noindent\textbfit{(iii) Distributed and Hierarchical Multi-Model Collaboration}
Edge–cloud and multi-agent extensions broaden SD into large-scale collaboration. Fast edge–cloud SD~\cite{venkatesha2025fast}, SLED~\cite{li2025sled}, and heterogeneous SD frameworks~\cite{zhu2025efficient} distribute drafting to edge SLMs and verification to cloud LLMs using asynchronous batching. SpecServe~\cite{huang2025specserve} and CoSine~\cite{gao2025collaborative} add adaptive schedulers and confidence-prior verifiers for SLO-aware efficiency, while PICE~\cite{zhan2025pice} and CoS~\cite{fu2025fast} extend SD to semantic-level and multi-model collaboration through dynamic task routing and alternating proposer–verifier roles. Together, these systems turn SD into a distributed orchestration framework for scalable, low-cost LLM inference.

\paragraph{(3) Compute-optimal Test-time Scaling}
Test-Time Scaling (TTS) increases inference compute to improve performance, while compute-optimal TTS maximizes it under fixed budgets by selecting suitable model sizes and families, representing a form of SLM–LLM collaboration. BestRoute~\cite{ding2025bestroute} optimizes model routing and TTS sampling budgets to reduce cost, and AgentTTS~\cite{wang2025agenttts} extends this to multi-stage tasks through dynamic model and budget allocation.

\subsection{Discussion}
Existing cost-effectiveness collaborations demonstrate clear savings across compute, latency, communication, storage, and API costs by coordinating the ML lifecycle but remain limited by fragmented pipelines, inconsistent cost reporting, and brittle rules for determining when to hand off to the LLM~\cite{behera2025towards}. Future work should standardize end-to-end cost-effectiveness metrics and employ adaptive routing, paired with compression, bidirectional distillation, and related techniques, to ensure reliable savings across SLM–LLM collaboration paradigms.

\section{Privacy-preserving Cloud–Edge Collaboration}
\label{sec:collaboration_privacy}
Deploying LLMs on edge devices is limited by hardware constraints. A practical alternative is a cloud–edge framework where lightweight SLMs handle local, context-aware tasks, and cloud-based LLMs provide broader generalization \cite{lv2023duet,shao2025division}. However, transmitting sensitive data from trusted edge SLMs to semi-trusted cloud LLMs raises privacy concerns, as personally identifiable or proprietary information may be logged and leaked for training~\cite{duan_privacy_2024}, posing long-term data security risks.

To balance privacy and performance, privacy-preserving cloud-edge collaboration has emerged at both the inference and fine-tuning stages. At the \textit{inference stage}, SLMs act as {(1) Sensitive Information Gatekeepers}, filtering or anonymizing data before sending it to LLMs, or as {(2) All-Information Guardians}, retaining private context locally and integrating it after generation. At the \textit{fine-tuning stage}, collaboration occurs via {(1) SLMs as Learners} (learning from LLMs via distillation), {(2) LLMs as Learners} (adapting via proxy tuning using SLM signals), or {(3) Collaborative Learners} (jointly improving through federated knowledge transfer).

\subsection{Inference Stage}
\label{sec:privacy_inference}
\begin{figure}[b]
    \centering
    \begin{subfigure}[t]{0.99\linewidth}
        \centering
        \includegraphics[width=\linewidth]{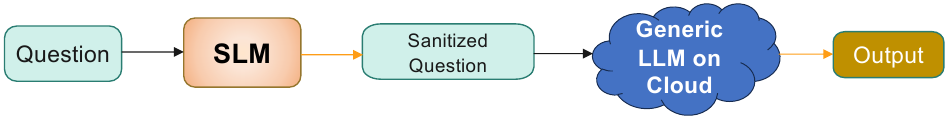}
        
        \caption{Sensitive information sanitization}
        \label{fig:gatekeeper1}
    \end{subfigure}
    \begin{subfigure}[t]{0.99\linewidth}
        \centering
        \includegraphics[width=\linewidth]{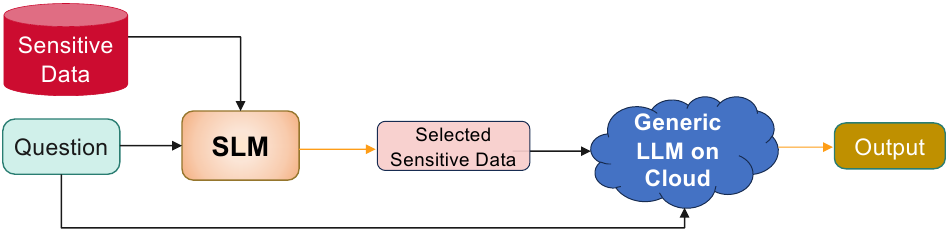}
        \caption{Selective disclosure}
        \label{fig:gatekeeper2}
    \end{subfigure}%
    \caption{SLMs as Sensitive-Information Gatekeepers.}
    \label{fig:gatekeeper}
\end{figure}

\begin{figure}[t]
    \centering
    \includegraphics[width=0.99\linewidth]{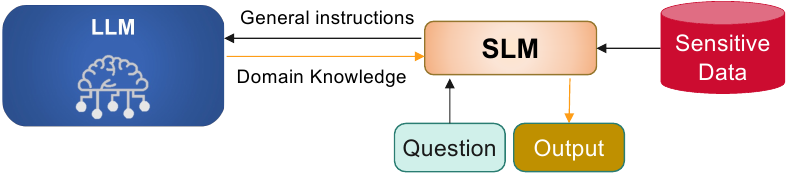}
    \caption{SLMs as All-Information Guardians.}
    \label{fig:privacy_guardian}
\end{figure}

During inference, edge user requests may contain sensitive information. Edge SLMs can process locally but with lower quality than cloud LLMs, while direct cloud LLM use risks data exposure. Thus, privacy-preserving edge–cloud collaboration is needed, based on protection levels, categorized into (1) Sensitive Information Gatekeepers (Fig.~\ref{fig:gatekeeper}) and (2) All-Information Guardians (Fig.~\ref{fig:privacy_guardian}).


\paragraph{(1) SLMs as Sensitive Information Gatekeepers} 
To protect sensitive information in edge–cloud collaboration, the edge SLM serves as a privacy-preserving filter, for (i) sanitizing sensitive content and (ii) selectively disclosing necessary information before transmitting requests to the cloud.

\noindent\textbf{\textit{(i) Sanitizing sensitive information}:}
Fig.~\ref{fig:gatekeeper}(a) shows how SLMs remove or rephrase prompts containing personally identifiable information (PII). Casper \cite{chong_casper_2024} use an SLM to filter PII before cloud transmission, while \citet{papadopoulou_neural_2022} employ SLM probabilities for detection. LLM
gatekeeper \cite{uzor_guarding_2025} rephrases queries to preserve meaning while removing PII, applied to clinical data~\cite{wiest_anonymizing_2024}. \citet{hartmann-etal-2024-llms} enhance SLMs by masking sensitive data and using LLM-generated examples.

\noindent\textbf{\textit{(ii) Selective disclosure}:}
Fig.~\ref{fig:gatekeeper}(b) illustrates selective disclosure, where the edge SLM shares limited, task-relevant information to reduce privacy risks. CoGenesis \cite{zhang2024cogenesis} shares sketches or logits with the LLM, while CORE~\cite{fan2025core} ranks and filters relevant context. P$^3$Defer \cite{zhang2024privacy} employs a policy network to balance privacy and quality, and RemoteRAG \cite{cheng_remoterag_2024} enforces a differential privacy budget to mathematically bound potential leakage. In sum, these methods shift from full-text disclosure to abstracted representation sharing, mitigating data exposure in SLM–LLM collaboration.


\paragraph{(2) SLMs as All-Information Guardians}
Edge device policies may impose strict privacy requirements that prohibit any PII disclosure. Under such constraints, edge SLMs act as \textit{All-Information Guardians}, processing sensitive data entirely on-device while integrating cloud LLM knowledge, ensuring that private information never leaves the user’s control, as shown in Fig.~\ref{fig:privacy_guardian}.
Sketch-based CoGenesis~\cite{zhang2024cogenesis} exemplifies the paradigm, using the cloud LLM for high-level planning while the SLM personalizes results locally with user logs. PrivacyBoost-SLM \cite{zhang_enhancing_2024} applies this to clinical decision-making, combining cloud-derived medical knowledge with patient data. RemoteRAG \cite{cheng_remoterag_2024} extends this idea by pairing SLMs with local retrievers (RAG) to integrate private and domain-general knowledge without exposing raw data. \citet{hartmann-etal-2024-llms} further employ LLMs as RAG example providers to enhance the SLM context. Industrial systems such as Apple Intelligence~\cite{apple_foundation_models} adopt a similar approach, processing sensitive data on-device and sending only abstracted information to the cloud.

\subsection{Fine-tuning Stage}
\label{sec:privacy_train}
\begin{figure}[t]
    \centering
    \vskip -0em
    \includegraphics[width=0.99\linewidth]{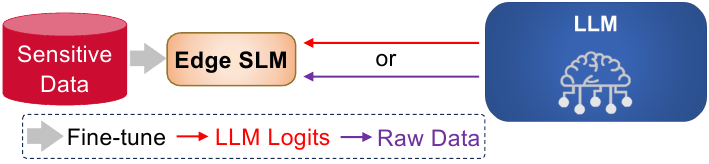}
    \caption{SLMs as Learners.}
    \label{fig:slm_as_learners}
\end{figure}

\begin{figure}[b]
    \centering
    \includegraphics[width=0.99\linewidth]{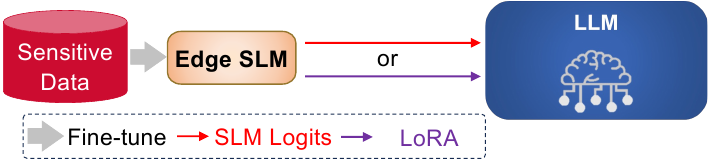}
    \caption{LLMs as Learners.}
    \label{fig:privacy_slms_as_learners}
\end{figure}

\begin{figure}[t]
    \centering
    \vskip -0em
    \includegraphics[width=0.99\linewidth]{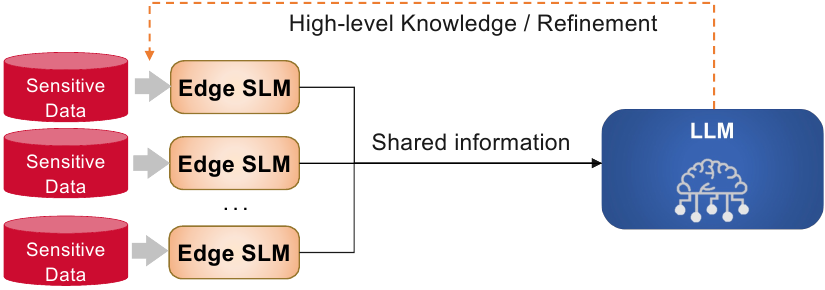}
    \caption{Cloud-Edge Collaborative Learning.}
    \label{fig:privacy_federated}
\end{figure}

During fine-tuning, edge data can facilitate the evolution of both edge and cloud LLMs by providing valuable fine-tuning signals. However, ensuring data protection without exposing sensitive information to the cloud remains a major challenge. To address this, privacy-preserving edge–cloud training frameworks have been proposed, which can be categorized into three paradigms based on the learning target:
(1) SLMs as Learners: the edge SLM acquires new capabilities from larger LLMs through local fine-tuning, distillation, or personalization, guided by cloud LLMs without revealing raw data (Fig.~\ref{fig:slm_as_learners});
(2) LLMs as Learners: the cloud LLM improves by consuming proxy signals or adapters distilled from locally trained SLMs, thereby enhancing generalization without accessing private data (Fig.~\ref{fig:privacy_slms_as_learners}); and
(3) Collaborative Learners: both SLMs and LLMs co-evolve through federated or privacy-preserving coordination (Fig.~\ref{fig:privacy_federated}).
\paragraph{(1) SLMs as Learners} 
Fig.~\ref{fig:slm_as_learners} shows edge SLMs learning from cloud LLMs to improve their ability and reduce cloud dependence through subsequent local fine-tuning. SLMs can learn via two modes: (i) raw text, such as reasoning traces or samples, and (ii) feedback logits as supervision signals.

\noindent\textbf{\textit{(i) Raw Data}:}
SLMs enhance reasoning by learning from cloud LLM outputs while keeping private data local. The LLM provides interpretable supervision (e.g., responses, reasoning traces) or synthetic samples for SLM fine-tuning. MiniLLM~\cite{gu2023minillm} and LlamaDuo~\cite{park-etal-2025-llamaduo} distill reasoning traces, while DRAG~\cite{chen2025dragdistillingragslms} extends to retrieval-augmented reasoning. Privacy-aware frameworks such as HomeLLaMA~\cite{homellama2025} and ~\citet{qin2024enabling} use selective textual supervision to align SLMs with user preferences under strict privacy constraints.

\noindent\textbf{\textit{(ii) Logit-based Learning}:}
SLMs learn from teacher LLMs’ numerical outputs, using logits as soft targets to imitate their decision boundaries without accessing raw text. TinyLLM~\cite{tianAnswers2024a} distills from multiple teachers for robustness, Mix Distillation~\cite{liSmall2025} combines LLM and SLM supervision, and ADPA~\cite{gaoAdvantageGuided2025} introduces an advantage-based objective for better reasoning alignment, ensuring privacy and efficiency through abstracted signal exchange.



\paragraph{(2) LLMs as Learners}
LLMs can be adapted to sensitive data without direct access by leveraging an edge SLM tuned locally. The SLM forwards only supervision signals, such as (i) logits or (ii) LoRA signals, to the LLM, enabling privacy-preserving alignment, as shown in Fig.~\ref{fig:privacy_slms_as_learners}.

\noindent\textbf{\textit{(i) Logit-based Learning}:}
Instead of directly fine-tuning an LLM, \citet{ormazabalCombLMAdaptingBlackBox2023} demonstrate that an LLM can be steered by tuning an SLM on private data and exposing only the SLM’s output logits to guide the LLM’s responses. 
Specifically, they train a lightweight MLP on the logits produced by both the SLM and the LLM, which is then used to steer the LLM's output during inference. 
This approach enables the use of private data for adaptation while avoiding its direct exposure to the LLM. An extension of this method involves applying direct logit offsets between the tuned model and the LLM~\cite{liu2024tuning,prada2025}, 
by computing the logits difference between the trained SLM and its frozen counterpart. The LLM adds this offset to its own output logits during inference. 

Follow-up work has examined the limitations of this approach. CPT~\cite{heCPTConsistentProxy2024} and GradOT~\cite{yaoGradOTTrainingfreeGradientpreserving2025} address inconsistencies arising from discrepancies between training and inference pipelines. Whereas the SLM is trained alone, it is expected to interact with its frozen counterpart to compute the logits offset and steer the LLM during inference. To address this gap and improve alignment,
CPT integrates both the frozen SLM and LLM in the training process to closely mirror the inference phase interaction and include the LLM feedback in the gradient updates. On the other hand, GradOT approximates the gradient updates using offsite-tuning~\cite{xiao2023offsitetuningtransferlearningmodel} to enable efficient alignment without co-training. To enhance robustness, GOOD~\cite{fangGOODDecodingTimeBlackBox2024} decodes the SLM’s output into text and re-encodes it for the LLM to extend logit steering to SLM–LLM pairs without a shared vocabulary. 

\noindent\textbf{\textit{(ii) LoRA-based Learning}:}
Recent work leverages LoRA signals to propagate knowledge instead of requiring logit outputs from SLMs to enable adaptation without direct access to private data. 
LoRASuite~\cite{liLoRASuiteEfficientLoRA2025} and LoRA-X~\cite{farhadzadehLoRAXBridgingFoundation2025} demonstrate how LoRA signal transfer can be applied across model versions and architectures. Similarly, Trans-LoRA~\cite{wangtextitTransLoRA2024} and Cross-LoRA~\cite{xiaCrossLoRADataFreeLoRA2025} extend this capability to heterogeneous model families, enabling more generalizable and privacy-conscious adaptation of LLMs.


\paragraph{(3) SLMs and LLMs as Collaborative Learners}
Beyond one-way transfer, collaboration can be bidirectional: SLMs and LLMs co-adapt via shared representations, logits, or adapter weights while preserving privacy. Fig.~\ref{fig:privacy_federated} shows multiple edge SLMs learn from local private data and transmit abstracted updates, such as logits or LoRA adapters, to the cloud LLM for aggregation and refinement, and then the cloud LLM returns high-level knowledge or refinement feedback to the local.
%
%
CrossLM~\cite{deng2025crosslm} achieves mutual enhancement by combining private task data and LLM-generated samples for privacy-preserving co-training. LSRP~\cite{lsrp2025} enables bidirectional feedback, where the LLM guides SLM retrieval and refines its own responses. Federated frameworks such as FedMKT~\cite{fan-etal-2025-fedmkt}, FedPT~\cite{gaoFedPTFederatedProxyTuning2024}, FedPFT~\cite{pengFedPFT2024}, FDLoRA~\cite{qiFDLoRAPersonalizedFederated2024}, and the Automated Federated PEFT Pipeline~\cite{fangAutomatedFederatedPipeline2024} aggregate client-side SLM adapters or logits to update central LLMs while preserving data privacy. FLoRA~\cite{wangFLoRA2024} further improves personalization through enhanced edge aggregation.
\subsection{Discussion}
Current cloud–edge SLM–LLM collaborations face challenges in achieving secure peer-to-peer exchange, effective cold-start learning, formal privacy guarantees, and scalable personalization under strict data protection. Future work should develop adaptive frameworks enabling privacy-preserving collaboration and lightweight personalization for policy-compliant intelligence in sensitive domains such as healthcare and customer service.




\section{Collaboration for Trustworthiness}
\label{sec:collaboration_trust}

LLMs offer strong generative ability and broad knowledge coverage but face major trustworthiness challenges in real-world use. Uncontrolled training data can lead to hallucinations, harmful or biased outputs, privacy leaks, and vulnerabilities exploitable through jailbreaks or adversarial attacks \cite{zhang2025catastrophic,liu2023maximum,wu2025image,zhang2025which,liu2020co,wang2021macrobert}. Updating LLMs for safety is costly and may harm generation quality. A more practical solution is integrating \textit{safety policies} that guide generation without retraining. SLMs serve this role effectively due to their compactness, independence from LLM architectures, and ease of policy updates. Recent SLM–LLM collaborations use SLMs to encode or enforce safety policies while LLMs handle content generation. Two paradigms dominate: (1) \textit{Safety-guided decoding}, where an SLM modifies LLM logits during decoding for token-level control; and (2) \textit{Guardian–generator}, where the SLMs trained as Guardians filter inputs and audit outputs, providing interaction-level governance. These paradigms are shown in Figs.~\ref{fig:trust_decoding} and~\ref{fig:trust_guard}.




\subsection{Safety-Guided Decoding}
\label{sec:safety_guided_decoding}
\begin{figure}[t]
    \centering
    \vskip -0em
    \includegraphics[width=0.99\linewidth]{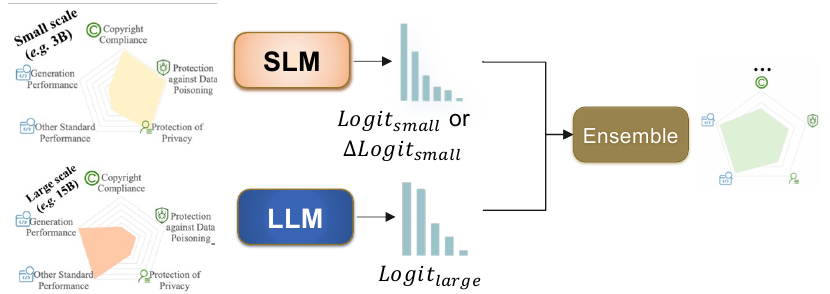}
    \caption{SLM Guided Safe LLM Decoding.}
    \label{fig:trust_decoding}
\end{figure}

Unsafe outputs in LLMs often arise during decoding; thus, integrating safety policies into token-level generation is crucial yet challenging, as it is difficult to anticipate which tokens may trigger unreliable content. Recent studies employ safety-tuned SLMs, fine-tuned for risks such as jailbreaks, privacy leaks, or bias, to guide LLM decoding in a safety-aware manner. As shown in Fig.~\ref{fig:trust_decoding}, SLM and LLM logits are fused to ensure both quality and safety through two strategies: (1) direct logit fusion (linear combination) and (2) logit offset fusion (using SLM logits as offsets to steer LLM probabilities toward safer tokens).

\paragraph{(1) Direct Logit Fusion}  
The CP-$\Delta$ theory~\cite{pmlr-v202-vyas23b} provides an information-theoretic foundation for model fusion, demonstrating that combining the predictive distributions of two models, $q_1(y|x)$ and $q_2(y|x)$, with distinct copyright risks yields a fused distribution
$p(y|x) \propto q_1(y|x)^{\alpha} q_2(y|x)^{1-\alpha}$ that minimizes weighted KL divergence and balances their strengths without retraining, where $\alpha$ controls each model’s contribution. The fused distribution lies closer to both sources than they are to each other, thus achieving a purification effect, retaining desirable properties while reducing copyrighted content. This operation is equivalent to linear weighting in logit space, $z_p = \alpha z_1 + (1-\alpha) z_2$, providing the theoretical foundation for SLM–LLM collaborative logit fusion. Subsequent methods build on this principle: Purifying LLMs~\cite{li2024purifying} fuses clean SLM and contaminated LLM logits to mitigate copyright, privacy, and backdoor risks, while MOD~\cite{shi2024decoding} extends fusion to multiple objectives, integrating models optimized for fluency, faithfulness, and safety to enable controllable generation.


\paragraph{(2) Logit Offset Fusion} 
\citet{mitchell2024an} introduced the concept of logit offset fusion by decomposing a fine-tuned model’s output into the base model’s logits and a behavioral delta: 
\begin{equation}
z_{\text{ft}}(x) = z_{\text{base}}(x) + \Delta(x),
\end{equation}
where $\Delta(x)$ represents the \textit{behavioral delta} induced by fine-tuning. 
Through an \textit{up-scaling} strategy, it integrates a large pre-trained model’s logits $z_{\text{large}}$ with the behavioral delta of a small fine-tuned model $\Delta(x) = z_{\text{ft}}(x) - z_{\text{small}}(x)$:
\begin{equation}
z_{\text{fusion}}(x) = z_{\text{large}}(x) + \lambda \, \Delta(x),
\end{equation}
where $\lambda$ is a scaling factor aligning the magnitude of the small model’s delta with the large model’s logit scale. This simple logit-level combination achieves up to 73\% of the full large-model fine-tuning helpfulness without re-training the large model.
This demonstrates that safety-aligned small models’ behavioral deltas can enhance LLM safety when fused at the logit level. Building on this idea, Offset Unlearning~\cite{huang2025offset} trains SLMs to generate targeted logit offsets that suppress sensitive or fabricated knowledge for privacy-preserving unlearning. Conversely, Weak-to-Strong Jailbreak~\cite{zhao2025weaktostrong} reveals that unsafe SLMs can steer safe LLMs toward harmful outputs, implying that the same mechanism can defensively employ safe SLMs to counter jailbreak attacks through safety-guided logit fusion.


\subsection{SLM as Guardian, LLM as Generator}
\label{sec:guardian}
\begin{figure}[t]
    \centering
    \vskip -0em
    \includegraphics[width=0.99\linewidth]{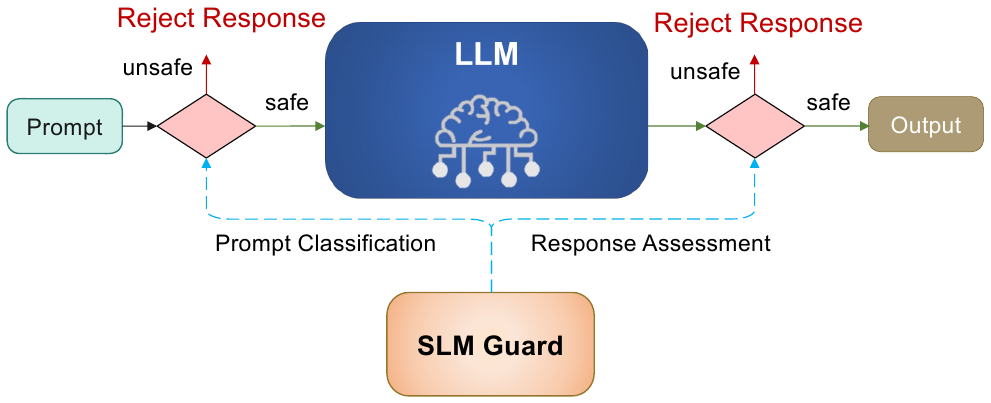}
    \caption{SLM Guardian, LLM Generator.}
    \label{fig:trust_guard}
\end{figure}

Guardrail-based approaches enhance LLM safety by placing SLMs before and after generation. In the SLM-as-Guardian and LLM-as-Generator paradigm, as shown in Fig.~\ref{fig:trust_guard}, SLMs, encoding the safety policy, handle input filtering and output auditing, offering configurable, interpretable, and reusable safety control across evolving applications.
Building on this paradigm, later studies advanced the framework. Llama Guard~\cite{inan2023llama} built a bidirectional safety pipeline using a fine-tuned LLaMA-2-7B classifier. Llama Guard 2~\cite{metallamaguard2} added multi-label risk classification; Llama Guard 3~\cite{dubey2024llama} expanded multilingual support and inspired lightweight variants like Llama Guard 3-1B-INT4~\cite{fedorov2024llama} and cross-modal Llama Guard 3 Vision~\cite{chi2024llama}. Llama Guard 4~\cite{llama_guard_4} further refined the taxonomy to 14 safety categories and was adapted for healthcare auditing~\cite{gangavarapu2024enhancing}. For lightweight protection, Prompt Guard~\cite{meta2024promptguard} and Prompt Guard 2~\cite{llama-prompt-guard2}, based on DeBERTa-v2 \cite{he2021deberta}, serve as fast, low-cost input filters. ShieldGemma~\cite{zeng2024shieldgemma, zeng2025shieldgemma} and WildGuard~\cite{han2024wildguard} use synthesized, large safety data to boost robustness against adversarial prompts. ThinkGuard~\cite{wen-etal-2025-thinkguard} adds ``slow thinking'' for better interpretability and defense. Beyond text safety, MiniCheck~\cite{tang2024minicheck} applies the guardian concept to factual verification, while Cascade Guardrails~\cite{nagireddy2024doubt} explores hierarchical defenses. LlamaFirewall~\cite{chennabasappa2025llamafirewall} extends this paradigm to multi-agent settings.

\subsection{Discussion} 
Current SLM–LLM collaborations for trustworthiness face challenges in unifying decoding and guarding, extending safety beyond text, enabling edge deployment, and standard evaluation. Future work should build integrated frameworks with multimodal safety, lightweight edge guardians, and standard evaluation benchmarks.

\section{Applications}
We summarize the representative works in Table~\ref{tab:slm_llm_overall}, detailing their collaboration mechanisms, employed SLMs and LLMs, evaluation datasets, and goals. From these studies, we can summarize the applications under each category. 

\paragraph{Collaboration for performance} 
In guidance-generation, SLMs assist LLMs in reasoning and classification tasks, such as intent detection \cite{liang2024synergizing}, natural language inference \cite{xu2023small}, sentiment analysis \cite{mojarradi2024improving}, and multi-hop question answering \cite{lee2024can}. In division–fusion, they form multi-stage pipelines for domains such as clinical analysis \cite{gondara2025elm}, factual verification \cite{hsu2024calm}, NL2SQL \cite{fan2024combining}, and trajectory modeling \cite{du2024trajagent}, where SLMs handle subtasks and LLMs integrate the results. 
This suggests that guidance-enhanced generation is more suitable for common general-domain tasks, where LLMs are the backbone generator and SLMs are trained on local downstream data, while division–fusion collaboration is preferred when both models exhibit comparable capabilities across different subtasks.

\paragraph{Collaboration for cost-effectiveness} 
The cost-effective SLM–LLM collaboration methods show stage-specific application preferences. 
Pre-training-stage frameworks target corpus construction and scaling efficiency on large datasets (e.g., Pile, FineWeb-Ed, C-Eval), supporting sustainable and scalable model development~\cite{rawat2024little,yano2025efficient,goyal2025distilled}.
Fine-tuning-stage methods focus on intent classification and reasoning transfer (e.g., BANKING, CLINC, MMLU, GSM8K), enabling smaller models to inherit LLM capabilities with minimal cost~\cite{jazbec2024efficient,zhang2025train,fan2025g}. 
Inference-stage approaches emphasize real-time reasoning and conversational tasks (e.g., AIME, GPQA, ShareGPT) where latency and compute efficiency are critical~\cite{chen2024frugalgpt,ong2024routellm,zhang2024privacy}. 

\paragraph{Collaboration for cloud-edge privacy} 
The inference-stage methods are particularly suited for end-user scenarios where privacy is a main concern. Edge SLMs prevent leakage of PII or sensitive context to external providers \cite{chong_casper_2024, zhang2024cogenesis}, supporting domains such as healthcare and customer service. 
Depending on privacy needs, {SLMs as Information Gatekeepers} may be applied \cite{zhang_enhancing_2024,cheng_remoterag_2024}. 
Meanwhile, {proxy tuning} enables domain-specific or personalized LLM adaptation without exposing private data, offering broader potential under edge resource privacy constraints and edge SLMs’ limited reasoning and context capacity~\cite{liLoRASuiteEfficientLoRA2025,prada2025,deng2025crosslm}.

\paragraph{Collaboration for trustworthiness} 
SLM–LLM collaboration for trustworthiness is applied in safety-critical and compliance-sensitive contexts. 
Safe decoding methods can correct hallucination, privacy, and toxicity risks in code generation, instruction following, and dialogue by steering the decoding distributions \cite{li2024purifying, li2024purifying, mitchell2024an,huang2025offset}. 
Guardian frameworks support content moderation, assistants, and fact verification by filtering directly unsafe inputs and ensuring output reliability \cite{inan2023llama, meta2024promptguard, zeng2024shieldgemma, wen-etal-2025-thinkguard}. 
This suggests that the former is better suited for applications requiring a balance between output quality and truthfulness, whereas the latter is preferred for highly secure applications that eliminate potentially risky inputs and outputs.

\section{Challenges and Future Directions}
\label{sec:future}

\noindent\textbf{Open and Interoperable SLM–LLM Ecosystem} Current SLM–LLM collaborations remain fragmented, with isolated models and task-specific systems limiting interoperability and reuse. Future research should develop open, cross-domain ecosystems that unify specialized and general models through standardized interfaces and shared repositories, enabling efficient model discovery, seamless composition, and collaborative benchmarking across diverse applications and domains.

\noindent\textbf{Benchmark Design for SLM–LLM Collaboration} Current SLM–LLM evaluations focus mainly on open-ended tasks and performance, overlooking structured tasks and system-level metrics. Future research should develop diverse benchmarks covering structured reasoning, modular cooperation, and system-level metrics, such as throughput, latency, and cost–performance, to better capture collaborative efficiency and real-world applicability.

\noindent\textbf{Synergistic SLMs as Decoders and Guardians.} Current safety mechanisms often separate decoding control from post-hoc moderation, limiting effectiveness in SLM–LLM safety collaboration. Future research should develop unified frameworks where SLMs function simultaneously as safety-aware decoders and guardians, enabling multi-stage, end-to-end safety governance that enhances trustworthiness across both generation and interaction phases.

\noindent\textbf{Cost-Aware SLM–LLM Collaboration} 
Current SLM–LLM collaborations emphasize performance over cost modeling, risking inefficient resource use. Future work should develop adaptive orchestration frameworks to optimize cost–performance trade-offs through predictive cost-aware routing, non-myopic collaboration policies considering long-term budgets, and multi-objective optimization balancing accuracy, latency, and financial cost for sustainable, scalable SLM–LLM systems.

\noindent\textbf{Theory for Privacy-Preserving Collaboration} Current privacy-preserving SLM–LLM collaborations lack formal theoretical foundations, relying mostly on empirical evaluations without quantifiable privacy assurances. Future research should establish rigorous metrics and provable bounds for data leakage and privacy preservation, enabling verifiable guarantees of data security and fostering trust in privacy-sensitive SLM–LLM systems.

\section{Conclusions}
In this paper, we present a systematic survey of SLM–LLM collaboration from the perspective of collaboration objectives: {performance enhancement}, {cost-effectiveness}, {cloud–edge privacy}, and {trustworthiness}. We analyze representative methods showing that most leverage SLMs’ lightweight adaptability.
We aim to highlight the current research state and provide insights for future work in this promising area of SLM–LLM collaboration.

\section*{Limitations}
Although our work introduces a systematic framework for analyzing SLM–LLM collaboration, several limitations remain. First, given the rapid evolution of language model technologies, certain implicit forms of SLM--LLM collaboration may not yet be fully captured, and further insights from the research community are encouraged to complement this study. 
Second, overlaps exist among several collaboration objectives, for instance, the personalization synergy, while not addressed in this survey, often intersects with cloud–edge collaboration, yet these cross-objective studies remain limited despite their conceptual and practical significance. 
We expect that addressing these challenges will motivate future comprehensive surveys and deeper investigations into SLM--LLM collaboration.

\bibliography{acl_latex}

\appendix

\end{document}